\documentclass[runningheads]{llncs}
\usepackage{graphicx}
\usepackage{bbding}
\usepackage[table,xcdraw]{xcolor}
\usepackage{hyperref}
\begin{document}
\title{Revisiting Data Scaling in Medical Image Segmentation via Topology-Aware Augmentation}
%

\author{Yuetan Chu\inst{1} \and
Zhongyi Han\inst{1} \and
Gongning Luo\inst{1(}\Envelope\inst{)} \and
Xin Gao\inst{1(}\Envelope\inst{)}}
\institute{King Abdullah University of Science and Technology (KAUST), Thuwal, Saudi Arabia\\
\email{{\{gongning.luo, xin.gao\}@kaust.edu.sa}}}


%
\maketitle              
%
\begin{abstract}
Understanding how segmentation performance scales with training data is fundamental for developing data-efficient medical AI systems. In this study, we systematically revisit data scaling behavior across 15 anatomical segmentation tasks spanning four imaging modalities. We observe that medical segmentation follows a structurally stable power-law-like relationship between predictive error and dataset size, characterized by rapid improvement in low-data regimes. However, unlike classical large-scale vision or language tasks, segmentation exhibits earlier and task-dependent performance saturation, with a persistent error floor emerging even as data increases. This behavior suggests that segmentation scaling is not purely data-constrained but is influenced by intrinsic geometric and anatomical structure. To further probe this geometry-constrained regime, we investigate whether topology-aware deformation-based augmentation can modify effective scaling dynamics. We compare random elastic deformation with registration-guided and generative deformation-field modeling strategies. While the overall functional form of the scaling law remains preserved, topology-aware augmentation systematically lowers the effective error scale and reshapes convergence behavior in a task-dependent manner, leading to improved sample efficiency without overturning the underlying scaling principle. These findings indicate that medical segmentation obeys a geometry-limited scaling law, and that anatomically grounded augmentation enhances data efficiency by expanding effective topological coverage rather than altering the fundamental scaling structure. Our results provide a principled empirical perspective on data-efficient learning in medical image segmentation. The code will be released at ***.

\end{abstract}

\vspace{-0.75em}
\section{Introduction}
\vspace{-0.75em}
Deep learning progress has been closely linked to neural scaling laws, which describe predictable performance improvements as model size, data volume, and computational resources increase\cite{scaling}\cite{power}. Such power-law behaviors have been widely observed across domains, including vision and language\cite{power_2}. However, their implications for image segmentation, particularly medical semantic segmentation, remain underexplored. Given the high annotation cost and clinical importance of medical image segmentation\cite{segmentation}\cite{anno}, understanding how performance scales with training data is critical for developing data-efficient and robust medical AI systems.

\begin{figure}[!t]
\centerline{\includegraphics[width=0.8\textwidth]{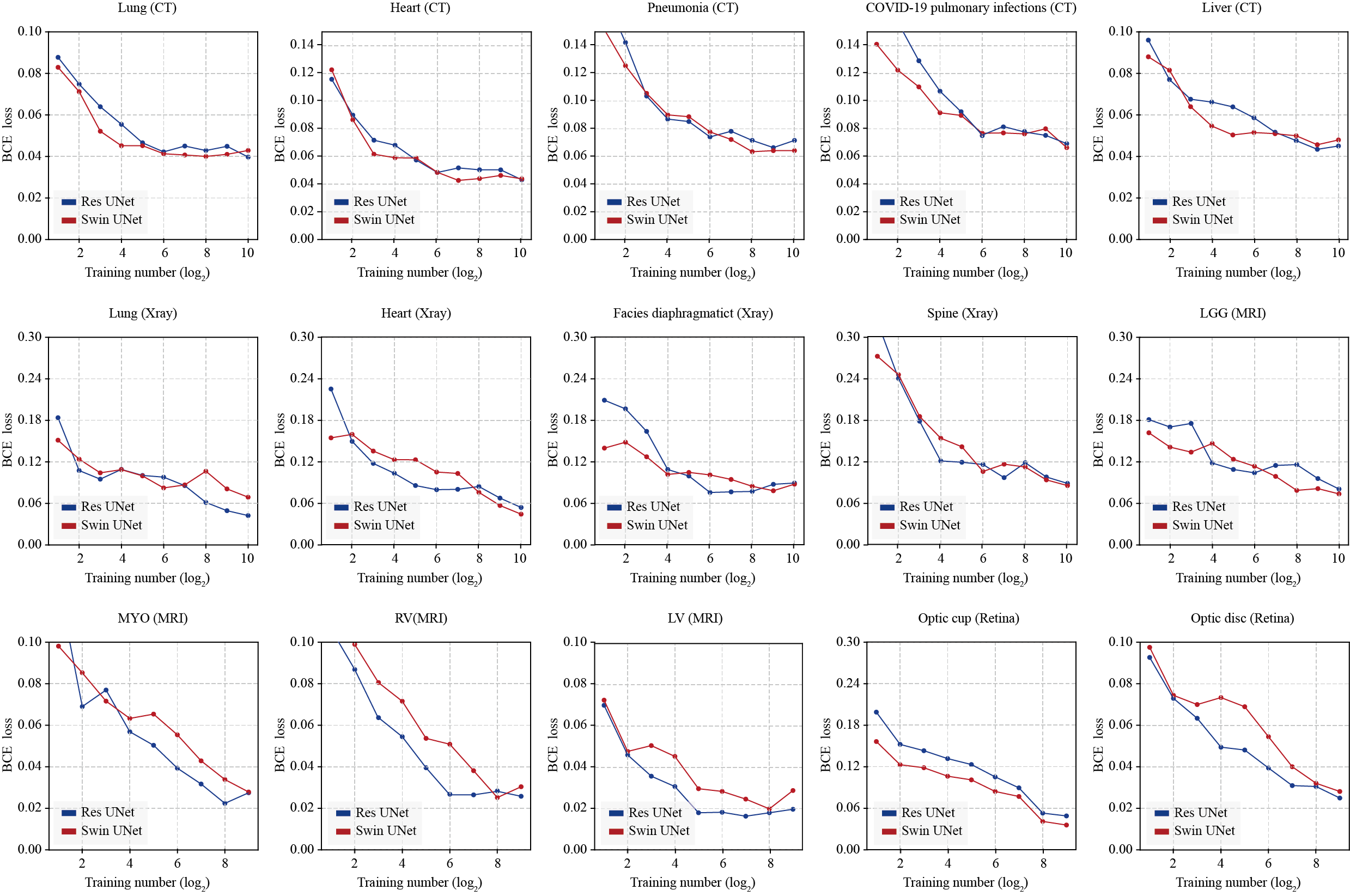}}
	\caption{Validation of data scaling law in medical segmentation tasks.}
 \vspace{-2.5em}
	\label{scaling}
\end{figure}

To better understand how segmentation performance scales with training data, we conduct a systematic empirical study across four imaging modalities, including X-ray, computed tomography (CT), magnetic resonance imaging (MRI), and retinal imaging, covering 15 anatomical segmentation tasks. By progressively increasing dataset size under a unified protocol, we observe consistent monotonic performance improvement across both convolutional architectures (nnUNet)\cite{segmentation} and transformer-based networks (Swin-UNet)\cite{swin} (Fig.~\ref{scaling}). In the low-data regime, error decreases rapidly and follows an approximate power-law-like trend. However, as dataset size grows, performance gains exhibit clear and task-dependent saturation, with persistent error floors emerging even before reaching large-scale data regimes. This early plateau suggests that medical segmentation scaling is not purely governed by data quantity, but is influenced by intrinsic geometric structure and constrained anatomical variability.

Human anatomy exhibits strong topological consistency across individuals, with organs sharing conserved geometric configurations despite inter-subject variation\cite{brain}\cite{regis_2}. If segmentation performance is indeed limited by effective geometric coverage rather than raw sample count alone, then augmentation strategies that expand anatomically plausible deformation manifolds should modify effective scaling behavior. To examine this hypothesis, we investigate deformation-based augmentation from a scaling perspective, including random elastic deformation, registration-guided diffeomorphic transformation\cite{LDDMM}\cite{brain}, and a learnable deformation-field generation approach, which is utilized to probe how altering topological coverage reshapes error decay dynamics. Empirically, topology-aware deformation strategies consistently improve performance in low-data regimes, suggesting that anatomical coverage, rather than dataset size alone, plays a critical role in determining segmentation efficiency.

\vspace{-0.5em}
\section{Scaling Law in Medical Segmentation}\label{scale}
\vspace{-0.75em}
In this section, we evaluate scaling behavior in medical segmentation using nnUNet (feature size 32, 1.96M parameters) and Swin-UNet (feature size 12, 1.59M parameters). Experiments span four imaging modalities and 15 tasks (Table~\ref{dataset}). For each task, the test set is fixed across scales, while training size increases exponentially (powers of two, subject to data availability). At each scale, training subsets are randomly sampled, models are independently re-initialized, and 20 trials are conducted. Models are trained with Adam (initial learning rate $1\times10^{-4}$) using cosine decay and 5-epoch warmup, with batch size 8. Inputs are resized to $512\times512$ and normalized to [0,1]. Baseline augmentation includes random flips, rotations ($\pm \pi/10$), affine transformations (scale 0.9-1.1, shear $\pm \pi/10$), and histogram shifting ($\pm0.2$). Training runs for up to 50 epochs with early stopping (patience 10) based on validation loss, retaining the best checkpoint. All experiments are conducted on a Linux workstation with an NVIDIA A6000 GPU.

For both training and evaluation, we used binary cross-entropy (BCE) as the sole optimization objective and error metric. During inference, no fixed thresholding (e.g., 0.5 binarization) was applied; instead, BCE was computed directly on probabilistic outputs against ground-truth masks. BCE corresponds to the negative log-likelihood under a pixel-wise Bernoulli model and is formally equivalent to cross-entropy used in classification tasks. This choice enables direct alignment with established neural scaling law literature, where predictive error is characterized in terms of cross-entropy or log-likelihood. Unlike bounded overlap-based metrics such as Dice and HD95, BCE provides a decomposable and information-theoretic measure of uncertainty, making it more suitable for analyzing continuous error decay and assessing whether segmentation performance follows a power-law relationship with respect to dataset size.

\begin{table*}[!h]
\vspace{-2.25em}
\centering
\resizebox{0.8\textwidth}{!}{
\begin{tabular}{|l|l|l|}
\hline
\rowcolor[HTML]{EFEFEF} 
Modality & Dataset Name                    & Segmentation tasks            \\ \hline
CT       & Private dataset                 & Lung, heart                   \\
CT       & Point-annotation segmentation\cite{anno} & Pneumonia                     \\
CT       & COVID-19 Seg Challenge\cite{covid}          & COVID-19 pulmonary infections \\
CT       & Abdomen-1K\cite{abdomen}                      & Liver                         \\ \hline
Xray     & Xray landmark\cite{landmark}                   & Lung, heart                   \\
Xray     & Torchxrayvision\cite{xtorch}                 & Facies diaphragmatica, spine   \\ \hline
MRI      & BraTS-TCGA-LGG\cite{lgg}                  & Low Grade Glioma (LGG)        \\
MRI      & ACDC\cite{acdc}                            & Left ventricle (LV), Right ventricle (RV), Myocardium (MYO) \\ \hline
Retinal (fundus)  & RiGA\cite{riga}\&PAPILA\cite{papila}                    & Optic disc, optic cup
\\ \hline      
\end{tabular}
}
\caption{Dataset details for segmentation scaling law validation.}
\label{dataset}
 \vspace{-2.5em}
\end{table*}

Fig.~\ref{scaling} illustrates the relationship between segmentation error and training set size across four imaging modalities and 15 tasks. For both nnUNet and Swin-UNet, BCE loss decreases monotonically and follows an approximately linear trend in log-scale over substantial regimes, consistent with power-law-like scaling. However, beyond the low-data regime, performance gains diminish and task-dependent saturation emerges, with persistent error floors appearing before extremely large data scales. Structurally constrained targets exhibit rapid early improvement followed by earlier flattening, whereas more heterogeneous tasks sustain improvement over broader ranges before reaching their limits. Importantly, these trends are consistent across convolutional and transformer-based architectures, indicating that the observed decay and saturation are intrinsic to the data–task geometry rather than model-specific effects. Overall, the baseline results suggest that medical segmentation follows a structurally stable scaling law with geometry-induced, task-dependent performance ceilings.

\vspace{-0.5em}
\section{Revisiting the Scaling Law from a Topological Perspective}
\vspace{-0.5em}
In this section, we examine the data scaling law when applying deformation-based augmentation approaches. We maintain all experimental settings consistent with the baseline configuration described in Section~\ref{scale}, while implementing these augmentation methods as external procedures.

\vspace{-0.5em}
\subsection{Random Elastic Deformation (RED)}
\vspace{-0.5em}
Random elastic deformation is a commonly used data augmentation technique that introduces non-linear spatial perturbations by sampling a coarse deformation grid and interpolating it to dense displacement fields~\cite{elastic}. In this study, we implemented elastic deformation using the MONAI framework~\cite{monai}. The grid spacing was set to $(10, 10)$ pixels with a deformation magnitude uniformly sampled from $[1, 3]$. Zero padding was applied at the image boundaries. The generated displacement field was applied to both the input image and its corresponding segmentation mask. Bilinear interpolation was used for image resampling, while nearest-neighbor interpolation was employed for segmentation masks. Elastic deformation was applied independently at each training iteration.

\vspace{-0.5em}
\subsection{Registration-Guided Deformation Augmentation (RegDA)}
\vspace{-0.5em}
Unlike random elastic deformation, RegDA generates augmentation fields via diffeomorphic image registration under the LDDMM framework~\cite{LDDMM}, where smooth and invertible transformations are parameterized by initial momenta and obtained through geodesic shooting.

Let $x$ be a training image and $\mathcal{Y}$ an external image set (n=50 in our experiments) strictly disjoint from all training, validation, and test data. No labels from $\mathcal{Y}$ are used; it serves solely to enrich anatomical deformation variability. For each pair $(x,y)$ with $y \in \mathcal{Y}$, we precompute the LDDMM registration and store the corresponding initial momentum $m_{x \to y}$. During training, for each $x$, three external images $\{y_i\}_{i=1}^{3} \subset \mathcal{Y}$ are randomly sampled, with associated momenta $\{m_{x \to y_i}\}_{i=1}^{3}$. Convex weights $(\lambda_1,\lambda_2,\lambda_3) \sim \mathrm{Dir}(1,1,1)$ are drawn to construct
\[
m_{\mathrm{comb}} = \sum_{i=1}^{3} \lambda_i \, m_{x \to y_i}.
\]
The deformation field is generated via
\[
\phi_{\mathrm{comb}} = \mathrm{Shoot}(m_{\mathrm{comb}}),
\]
which preserves the diffeomorphic structure since the combination occurs in momentum space. The transformation is applied to both the image and the label.
\[
(\hat{x}, \hat{s}) = (x, s) \circ \phi_{\mathrm{comb}},
\]
using bilinear interpolation for images and nearest-neighbor interpolation for masks. The weights are resampled at each iteration, yielding stochastic yet anatomically grounded augmentation.

\vspace{-0.5em}
\subsection{Generative Modeling of Deformation Fields (GenDA)}
\vspace{-0.5em}
While registration-guided augmentation enriches anatomical variability using external samples, its diversity is limited by the available datasets. To alleviate this constraint, we adopt a generative modeling strategy for deformation fields based on conditional adversarial learning, as described in \cite{genDA_1}\cite{genDA_2}

Let $\mathcal{W} = \mathcal{X} \cup \mathcal{Y}$ denote the union of the training set $\mathcal{X}$ and the external image set $\mathcal{Y}$. For ordered pairs $(w_i, w_j) \in \mathcal{W}$, we compute LDDMM registrations to obtain diffeomorphic transformations $\phi_{w_i \rightarrow w_j}$, represented as dense displacement fields
\[
u_{w_i \rightarrow w_j}(p) = \phi_{w_i \rightarrow w_j}(p) - p.
\]
These fields form a deformation dataset $\mathcal{U}$ used to train a conditional GAN (cGAN)~\cite{cgan}. The generator $G$ maps an image $w_i$ and noise $z$ to a displacement field $u_i^G = G(w_i, z)$, while the discriminator $D$ distinguishes real fields from $\mathcal{U}$ and generated samples conditioned on the same image.

To promote smoothness and discourage local folding, we incorporate regularization on the spatial Jacobian $\nabla u_i^G$, penalizing large gradients and negative Jacobian determinants. This encourages topology-preserving deformations, although diffeomorphism is not strictly guaranteed. Empirically, no substantial folding artifacts were observed. During augmentation, for a source image $x$, we sample $z$ to generate $u_x^G$, and scale it by $\mu \sim \mathrm{Uniform}(0,1)$ as $\tilde{u}_x = \mu \, u_x^G.$ The augmented pair is obtained as
\[
(\hat{x}, \hat{s}) = (x, s) \circ (\mathrm{Id} + \tilde{u}_x),
\]
using bilinear interpolation for images and nearest-neighbor interpolation for segmentation masks.

We emphasize that the external image set used in RegDA and GenDA contains no task-specific labels and serves solely to enrich anatomical deformation variability. While this introduces additional distributional information, it does not provide segmentation supervision. Our objective is to examine how expanding geometric coverage influences scaling behavior under controlled prior injection.

\vspace{-0.25em}
\subsection{Experimental results}
\vspace{-0.5em}

\begin{figure}[!t]
\centerline{\includegraphics[width=0.8\textwidth]{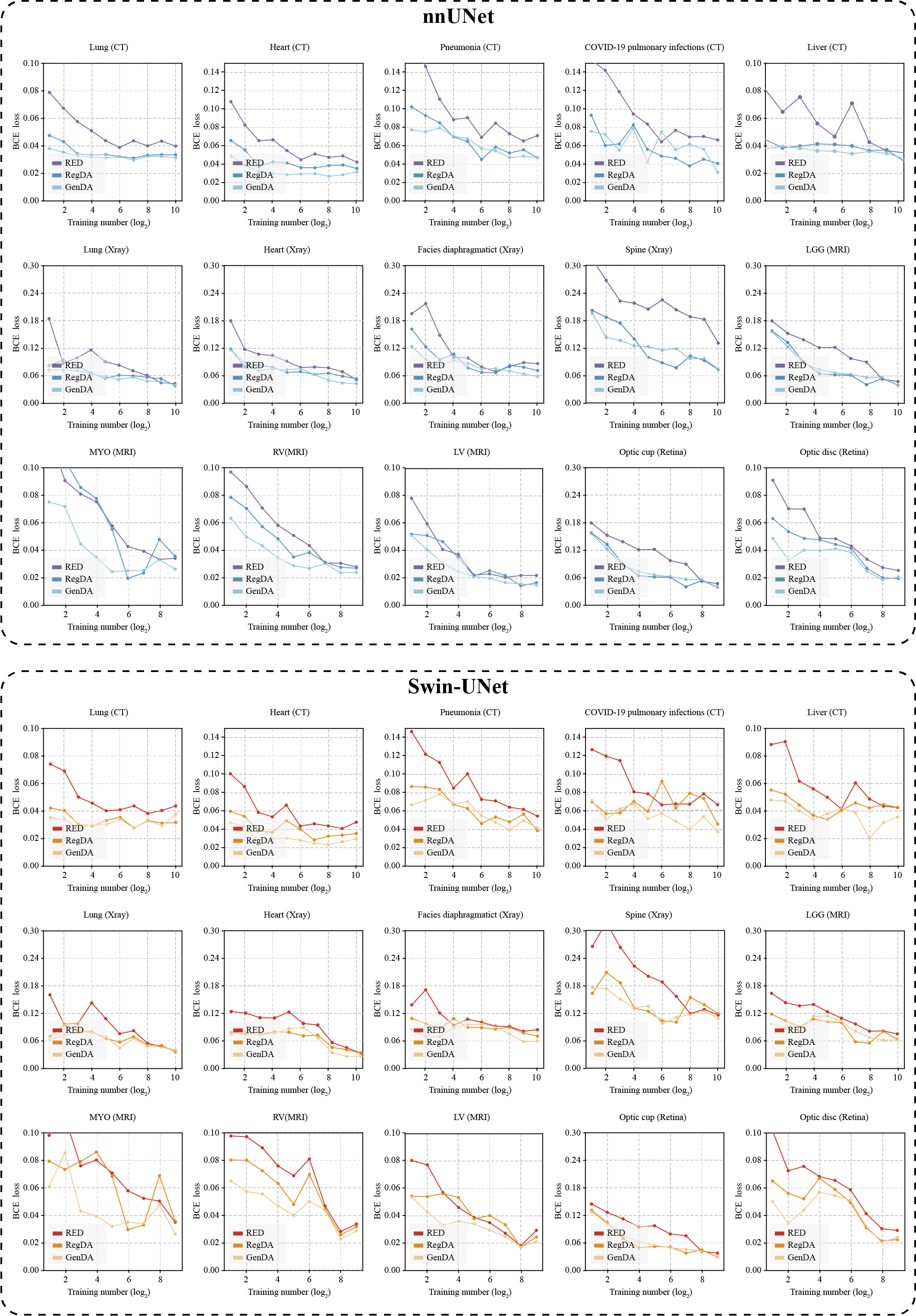}}
	\caption{Data scaling law comparison of different transformation techniques.}
\vspace{-2.5em}
\label{res}
\end{figure}

The results across nnUNet and Swin-UNet (Fig.~\ref{res}) show that topology-aware augmentation reshapes empirical scaling behavior while preserving its overall structure. Across methods, BCE loss maintains an approximately linear trend with respect to the logarithm of training size, indicating that the power-law-like functional form remains stable. Compared with RED, both RegDA and GenDA systematically lower the scaling curves, with the largest gains observed in the low-data regime (training size $< 2^4$), reflecting improved sample efficiency. As dataset size increases, performance gaps generally narrow, yet in several tasks the asymptotic error floor is also reduced. This indicates that augmentation predominantly enhances efficiency in the data-limited regime, while in certain cases it can also shift the attainable performance ceiling. In anatomically complex tasks, GenDA tends to yield more consistent improvements than RegDA, implying that richer deformation modeling better expands effective geometric coverage. Overall, these results indicate that topology-aware augmentation enhances efficiency by reshaping effective learning dynamics within a structurally stable scaling law, rather than altering the underlying scaling principle itself.

\section{Quantitative Scaling Law Fitting}
\vspace{-0.5em}

\begin{figure}[!t]
\centerline{\includegraphics[width=0.8\textwidth]{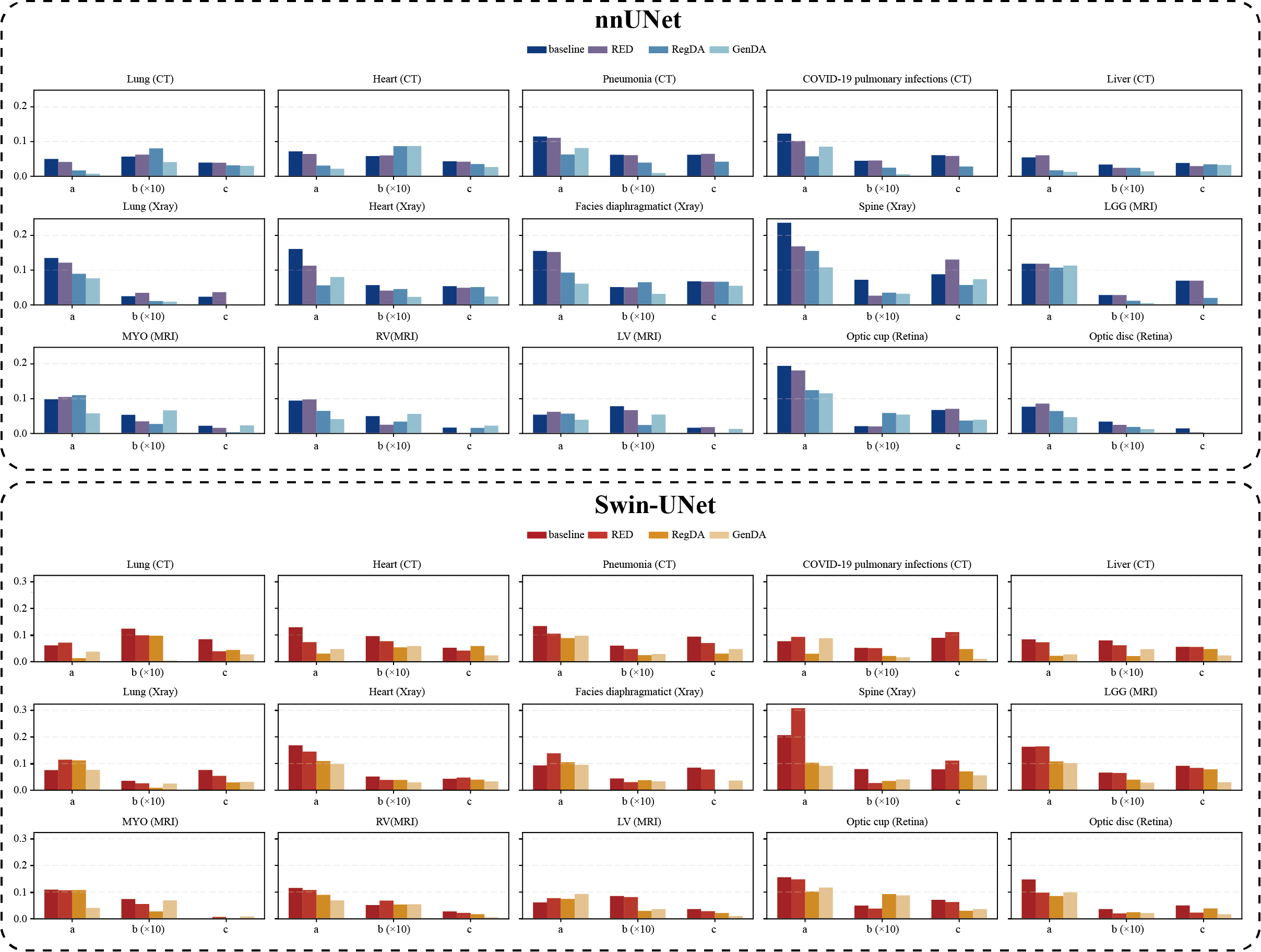}}
	\caption{Quantitative power-law fitting and parameter comparison across augmentation strategies.}
\vspace{-1.5em}
	\label{stat}
\end{figure}

To quantitatively characterize how segmentation error scales with training data size, we adopt a three-parameter power-law model with an irreducible error floor. Let $N$ denote the number of training samples and $E(N)$ the corresponding predictive error (measured by BCE). The scaling relationship is modeled as:
\begin{equation}
E(N) = a N^{-b} + c .
\end{equation}
Here, $a$ denotes the reducible error scale, controlling the magnitude of error decay in the low-data regime; larger values correspond to higher initial reducible error. The exponent $b$ captures the effective decay rate with respect to dataset size, with larger values indicating steeper reduction within the observed regime. The parameter $c$ represents the irreducible error floor, corresponding to the asymptotic limit as $N \to \infty$ and reflecting intrinsic task complexity and structural ambiguity. The model is fitted via nonlinear least squares across all scales for each task and architecture.

The quantitative results in Fig.~\ref{stat} show that, for baseline methods, the asymptotic floor $c$ remains non-negligible in most tasks, confirming that performance saturation is a common phenomenon in medical image segmentation. Absolute performance differences are primarily driven by data–task characteristics rather than architectural choice. Compared with the baseline, RED introduces only minor changes to the fitted parameters, indicating that random deformation provides limited improvement in effective data utilization.

In contrast, topology-aware augmentation (RegDA and GenDA) preserves the overall functional form of the scaling relationship while modifying the fitted coefficients. Across most tasks, the coefficient $a$ is consistently reduced, corresponding to a systematic downward shift of the scaling curve and improved efficiency in the low-data regime. Meanwhile, the scaling exponent $b$ exhibits substantial task-dependent variability without a consistent monotonic trend across methods, suggesting that augmentation reshapes the effective learning dynamics in a manner coupled to anatomical variability rather than uniformly accelerating convergence. In certain tasks, the asymptotic floor $c$ is also reduced, implying that topology-aware augmentation can, in some cases, influence not only early-stage efficiency but also the attainable error bound. Overall, these results indicate that segmentation scaling in medical imaging remains structurally stable, while topology-aware augmentation operates by modulating the effective error scale and, in a task-dependent manner, the decay dynamics, rather than altering the fundamental scaling law itself.

\vspace{-0.5em}
\section{Discussion and Conclusion}
\vspace{-0.75em}
In this study, we demonstrate that medical image segmentation obeys a structurally stable scaling law with geometry-induced saturation. Across diverse modalities and anatomical targets, segmentation error follows an approximate power-law-like decay as dataset size increases, yet exhibits persistent and task-dependent performance ceilings. These results indicate that segmentation scaling is not purely data-driven but is constrained by intrinsic geometric structure and limited anatomical variability. Topology-aware augmentation enhances sample efficiency by reshaping effective learning dynamics rather than overturning the scaling principle itself. Although the functional form of the scaling relationship remains preserved, augmentation modifies the fitted parameters in a task-dependent manner, improving low-data efficiency and, in some cases, lowering the attainable error floor. Notably, these effects are achieved without introducing additional supervision, but by incorporating unlabeled anatomical distributional information through deformation modeling. The fact that expanding geometric coverage alone can measurably alter scaling parameters further supports the interpretation that segmentation scaling is geometry-limited rather than purely annotation-limited.

Several limitations merit consideration. First, our analysis is restricted to moderate data scales; behavior in substantially larger regimes remains to be examined. Although the observed power-law-like trends are stable within the explored range, extrapolation beyond the studied scales should be interpreted cautiously. Second, experiments are conducted in 2D settings, and whether similar geometry-induced saturation persists in full 3D segmentation requires further study. Third, we focus on specific architectures and a single error metric (BCE); alternative modeling paradigms or evaluation criteria may reveal additional nuances. Finally, while deformation-based augmentation serves as a controlled probe of geometric effects, other forms of structural prior or data synthesis may interact with scaling dynamics differently, and our conclusions should therefore be understood as characterizing a principled yet partial view of segmentation scaling behavior.

\end{document}